\documentclass[10pt,conference]{IEEEtran}
\IEEEoverridecommandlockouts

\usepackage{cite}
\usepackage{amsmath}
\usepackage{amsfonts}
\usepackage{bm}
\usepackage{graphicx}
\usepackage[bookmarks=false]{hyperref}
\usepackage{amsthm}
\usepackage{amssymb, subcaption, url, array, booktabs, xcolor}
\usepackage{threeparttable}
\usepackage{multirow}
\usepackage{pifont}

\DeclareMathOperator*{\argmax}{arg\,max}

\begin{document}

\title{Robust Beam Prediction for V2X Networks with Multi-Modal Sensing}

\author{
\IEEEauthorblockN{Chen Shang$^{1}$, Dinh Thai Hoang$^{1}$, Diep N. Nguyen$^{1}$, and Jiadong Yu$^{2}$}
\IEEEauthorblockA{$^{1}$School of Electrical and Data Engineering, University of Technology Sydney, Australia}
\IEEEauthorblockA{$^{2}$Internet of Things Thrust, The Hong Kong University of Science and Technology (Guangzhou), Guangzhou, China}
% \IEEEauthorblockA{Email: chen.shang@student.uts.edu.au, \{hoang.dinh, diep.nguyen\}@uts.edu.au, jiadongyu@hkust-gz.edu.cn}
}

\maketitle
\begin{abstract}
Integrated sensing and communication (ISAC) provides a promising foundation for beam prediction in future vehicle-to-everything (V2X) networks. However, existing sensing-assisted beamforming methods still rely heavily on radio-frequency sensing, which may become unreliable in complex vehicular environments. Meanwhile, the growing availability of heterogeneous sensors, such as cameras and LiDAR, offers new opportunities to improve beam prediction through richer environmental perception. Motivated by this, this paper proposes a multi-modal beam prediction framework for V2X networks. Specifically, we develop BeamTransFuser, a hierarchical Transformer-based architecture that progressively fuses camera, LiDAR, radar, and GPS observations for robust beam prediction. In addition, to handle possible missing modalities in practical deployment, we introduce a generative module that reconstructs missing modality features from the available observations. Experimental results on a real-world multi-modal V2X dataset show that the proposed framework consistently outperforms representative baselines, while the generative module further improves robustness under incomplete sensing conditions.
\end{abstract}


\vspace{-15pt}
\section{Introduction}
Wireless communication systems have evolved steadily over the past decades toward more intelligent and efficient operation. Among the technologies envisioned for sixth-generation (6G) networks, integrated sensing and communication (ISAC) has emerged as a particularly important paradigm~\cite{ITU2023}. By unifying sensing and communication within the same framework, ISAC can lower signaling overhead, reduce beam alignment delay, and improve spectral efficiency~\cite{11358925}. These advantages are especially relevant to vehicle-to-everything (V2X) networks, where rapid mobility and fast-varying link conditions require highly accurate beam alignment to sustain high-data-rate and low-latency communication services~\cite{11091493,11603558}.

Despite its promise, practical ISAC deployment in vehicular environments still faces significant challenges. On the one hand, existing ISAC design paradigms generally involve inherent trade-offs among sensing accuracy, communication efficiency, and implementation complexity~\cite{10147248,11554053}.
Specifically, sensing-centric designs usually prioritize sensing performance but may sacrifice communication throughput and standard compatibility, whereas communication-centric designs focus on transmission efficiency at the expense of sensing capability. As a result, neither of these two designs can simultaneously provide strong sensing and communication performance in a balanced manner. On the other hand, joint-design approaches attempt to balance both functions, yet they often require more sophisticated signal processing, tighter synchronization, and higher hardware overhead, which makes practical real-time implementation more challenging~\cite{10147248}.

Furthermore, a common characteristic of these approaches is that they still rely predominantly on radio-frequency (RF)-based sensing. In dense urban roads, the quality of RF observations can deteriorate significantly because of blockage, multipath effects, and non-line-of-sight (NLoS) transmission conditions~\cite{10944644}.
Such impairments may degrade sensing fidelity, distort environmental perception, and eventually reduce beam alignment reliability. This issue becomes even more critical in V2X networks, where high vehicle mobility and rapidly changing road conditions place stricter requirements on accurate and timely beam prediction. Therefore, relying on RF sensing alone is often insufficient for practical and reliable beam prediction in realistic V2X scenarios.

In this context, exploiting heterogeneous sensing modalities beyond RF signals becomes a natural way to improve beam prediction robustness~\cite{10330577}. For example, the position-based method in~\cite{10683225} relies mainly on GPS information for beam prediction, while TII~\cite{tian2023multimodal} incorporates camera and GPS data through a Transformer-based design. CMDF~\cite{10.1145/3653644.3680497} and ICMFE~\cite{10912462} further exploit camera-radar sensing for beamforming-related tasks. Moreover, the multi-modal method in~\cite{10683225} considers richer sensing combinations involving camera, LiDAR, radar, and GPS. These works verify that introducing additional sensing modalities can improve beam prediction by providing complementary geometric, semantic, and environmental information. Nevertheless, most existing methods either consider only part of the available sensing modalities or lack a unified architecture that can fully exploit the complementarity among camera, LiDAR, radar, and GPS data. Moreover, they generally assume complete sensing inputs during inference. In practical deployment, however, some modalities may be unavailable due to sensor failure, environmental interference, or hardware constraints. In such cases, a model trained with fixed multi-modal inputs may encounter input mismatch and fail to operate properly during inference.

Motivated by the above observations, this paper develops a robust multi-modal beam prediction framework for practical V2X networks by jointly leveraging camera, LiDAR, radar, and GPS observations. Specifically, we propose BeamTransFuser, a hierarchical Transformer-based architecture that extracts modality-specific features and progressively fuses them through multi-stage cross-modal interaction, so as to better exploit the complementary information carried by these heterogeneous sensing modalities. In addition, to handle incomplete sensing conditions in practical deployment, we further introduce a generative module that reconstructs missing modality features from the available observations. In this way, the proposed framework can support effective beam prediction even when part of the sensing input is unavailable. The main contributions of this paper are summarized below:
\vspace{-1pt}
\begin{itemize}
    \item We propose BeamTransFuser, a hierarchical multi-modal beam prediction framework that jointly leverages camera, LiDAR, radar, and GPS observations for robust beam selection in V2X networks.
    \item We design a modality generation mechanism that reconstructs missing modality features from the available sensing inputs, thereby improving beam prediction robustness under incomplete sensing conditions.
    \item We conduct extensive experiments on a real-world multi-modal V2X dataset. The results demonstrate that the proposed framework consistently surpasses representative baseline methods in beam prediction performance, while the generative module effectively reconstructs missing modality features to enable reliable beam prediction under incomplete sensing conditions.
\end{itemize}

% The remainder of this paper is organized as follows. Section~\ref{system overview} presents the system model and problem formulation. Section~\ref{BeamTransFuser Framework} introduces the BeamTransFuser architecture. Section~\ref{sec:generation_only} describes the modality generation mechanism. Section~\ref{simultion} reports the performance evaluation results. Finally, Section~\ref{conclusion} concludes the paper.

\vspace{-4pt}
\section{System Overview and Problem Formulation}\label{system overview}
\begin{figure}[t]
    \centering
    \includegraphics[width=0.98\linewidth]{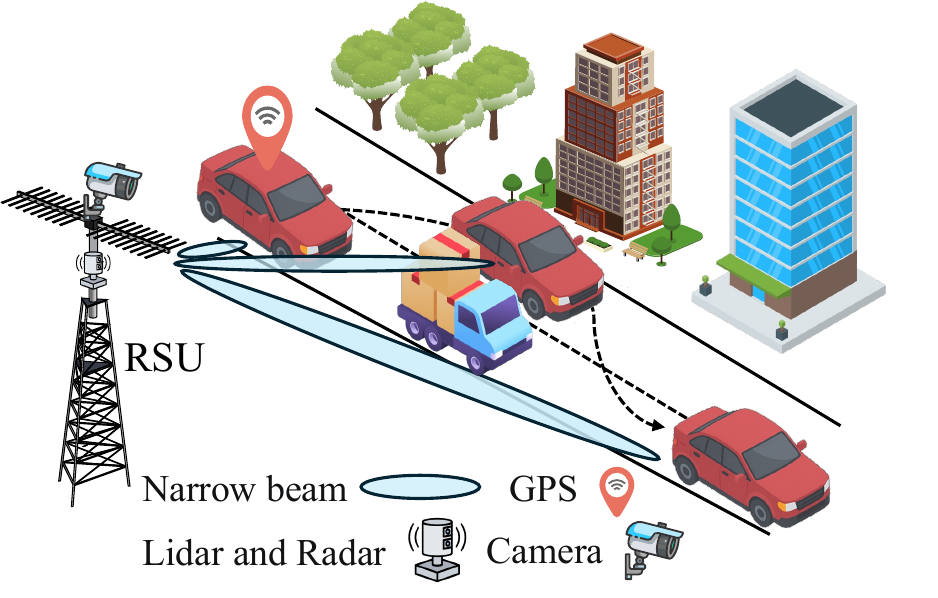}
    % \caption{Illustration of the considered multi-modal sensing-assisted beam prediction system. The overall procedure is: (1) the RSU-side sensors collect real-time observations, (2) the collected data are fed into a pretrained beam prediction model at the RSU, and (3) the RSU determines the beamforming decision according to the model output.}
    \caption{Multi-modal sensing-assisted beam prediction system. The RSU collects real-time observations, processes them with a pretrained model, and then determines the beamforming decision.}
    \label{fig:scenario}
\end{figure}

\vspace{-2pt}
Fig.~\ref{fig:scenario} presents the multi-modal sensing-assisted communication system considered in this work, where the RSU uses a uniform linear array for signal transmission and reception. Different from conventional ISAC frameworks that depend primarily on a single sensing source, the RSU integrates several sensing modalities, namely GPS, camera, LiDAR, and radar, to obtain real-time awareness of the surrounding environment.
Specifically, since these modalities provide different types of information, they can complement one another under diverse propagation and mobility conditions. For example, when RF sensing is degraded by blockage or NLoS propagation, LiDAR can still provide geometric structure information, while camera observations can offer useful semantic cues for beam prediction. As a result, the RSU can exploit these heterogeneous sensing inputs to support more reliable beam prediction under practical V2X conditions.

We consider a predefined beamforming codebook $\mathcal{F}=\{\mathbf{f}_m\}_{m=1}^{M}$ at the RSU, where each $\mathbf{f}_m$ denotes a complex-valued beamforming vector and $M$ is the total number of beam candidates. Let $\mathbf{f}_m$ be the transmit beamformer applied to the downlink symbol $s$. Accordingly, the received signal at the vehicle is given by:
\begin{equation}
\mathcal{C} = \mathbf{h}^{\mathrm{H}} \mathbf{f}_m s + z_c,
\label{eq:communication for vehicle}
\end{equation}
where $\mathbf{h}$ is the downlink channel vector between the RSU and the vehicle, and $z_c \sim \mathcal{CN}(0,\sigma^2)$ denotes circularly symmetric complex Gaussian noise. Accordingly, the signal-to-noise ratio (SNR) and the corresponding achievable rate can be respectively expressed as:
\begin{equation}
\gamma(\mathbf{f}_m)=\frac{|\mathbf{h}^{\mathrm{H}}\mathbf{f}_m|^2}{\sigma^2},~~R=\log_2\big(1+\gamma(\mathbf{f}_m)\big).
\label{eq:snr}
\end{equation}
% and the corresponding achievable rate can be expressed as:
% \begin{equation}
% R=\log_2\big(1+\gamma(\mathbf{f}_m)\big).
% \label{eq:rate}
% \end{equation}
Therefore, the optimal beamforming vector is selected as:
\begin{equation}
\mathbf{f}^{*}=\argmax_{\mathbf{f}_m\in\mathcal{F}} \log_2\big(1+\gamma(\mathbf{f}_m)\big).
\label{eq:optimal-beam}
\end{equation}

Instead of performing exhaustive beam search during online transmission, this work considers multi-modal sensing-assisted beam prediction, where heterogeneous sensing observations are directly used to infer the target beam index. Let $\mathcal{M}_{\boldsymbol{\Theta}}(\cdot)$ denote the beam prediction model parameterized by $\boldsymbol{\Theta}$. Suppose the training dataset is given by $\mathbf{\mathcal{X}}=\{\mathcal{X}_i\}_{i=1}^{N}$, where each sample contains multi-modal sensing inputs, and let $\mathbf{\mathcal{Y}}=\{\mathcal{Y}_i\}_{i=1}^{N}$ denote the corresponding beam-index labels. Therefore, the learning task can be formulated as:
\begin{equation}
\textbf{P1:}~ \boldsymbol{\Theta}^* = \underset{\boldsymbol{\Theta}}{\arg\min}~ \sum_{i=1}^{N} \mathcal{L} \left( \mathcal{M}_{\boldsymbol{\Theta}}(\mathcal{X}_i), \mathcal{Y}_i \right),
\label{eq:beamtransfuser}
\end{equation}
where $\mathcal{L}(\cdot)$ represents the training loss. The core difficulty of \textbf{P1} lies in how to effectively extract useful representations from heterogeneous sensing modalities and fuse them into a unified feature space for accurate beam prediction. To tackle this problem, we next introduce the proposed BeamTransFuser framework.

\section{The Proposed Multi-Modal Framework}\label{BeamTransFuser Framework}
\begin{figure}[t]
    \centering
    \includegraphics[width=0.98\linewidth]{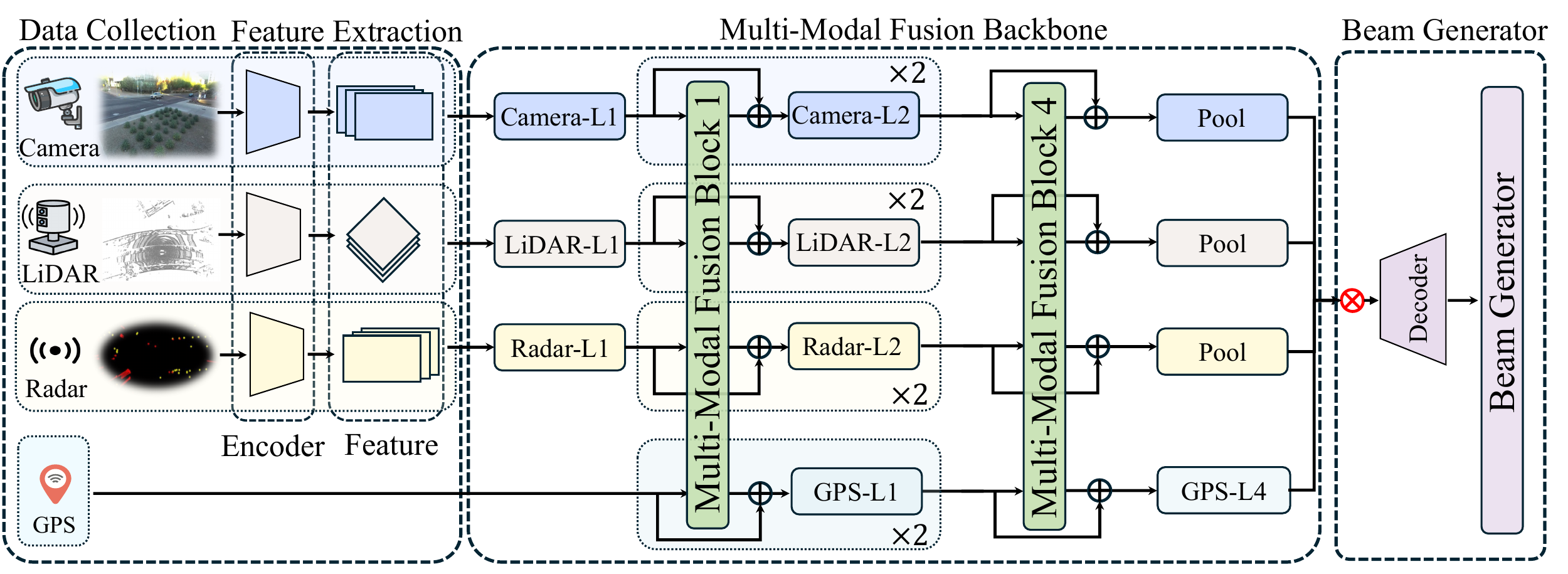}
    \caption{BeamTransFuser architecture with modality-specific branches and hierarchical cross-modal fusion.}
    \label{Fig.BeamTransFuser}
    % \vspace{-10pt}
\end{figure}

As shown in Fig.~\ref{Fig.BeamTransFuser}, the proposed BeamTransFuser consists of four modality-specific branches for camera, LiDAR, radar, and GPS inputs, followed by a hierarchical fusion backbone built on Transformer modules.
Specifically, each branch first extracts features from its corresponding sensing modality, allowing modality-aware representations to be learned before cross-modal interaction is introduced. The extracted features are then progressively fused through multi-modal fusion blocks inserted at different stages of the backbone, so that cross-modal dependencies can be modeled from low-level representations to high-level semantic features. Moreover, the backbone adopts a four-stage hierarchical design, where each modality is processed through progressive layers, while multi-modal fusion blocks are interleaved between adjacent stages to enable cross-modal interaction across different representation levels.
In addition, residual connections (i.e., $\oplus$) are employed to retain modality-specific information and improve training stability. After the final fusion stage, the resulting modality features are pooled and integrated by a learnable fusion module with softmax-normalized weights (i.e., ${\color{red}\otimes}$) before being passed to the decoder and beam generator for beam index prediction. The details of the proposed framework are presented as follows.

\vspace{-6pt}
\subsection{Multi-Modal Branches}
As shown in Fig.~\ref{Fig.BeamTransFuser}, camera, LiDAR, and radar inputs are first processed by dedicated convolutional encoders that transform raw data into spatial feature representations. In contrast, GPS data are low-dimensional and are mapped into the common feature space through a multilayer perceptron, without relying on the same spatial encoding pipeline. After this initial encoding step, each modality is further handled by a modality-aware feature extractor. Specifically, ResNet34~\cite{he2016deep} is used in the camera stream to extract discriminative visual features from RGB images, whereas lighter ResNet16~\cite{he2016deep} networks are adopted for the LiDAR and radar branches to reduce computational complexity while preserving essential structural cues.

To enable progressive cross-modal interaction, the encoded modality-specific representations are further processed by the hierarchical fusion backbone.
At each stage, the features are first aligned to a common representation scale and then converted into token sequences for Transformer-based fusion. This design allows modality-specific characteristics to be retained while gradually integrating complementary information across different sensing sources. Consequently, low-level geometric patterns and high-level semantic cues can be fused in a progressive manner, which is particularly useful in dynamic V2X scenarios where both local structural details and global contextual information influence beam prediction.

% \vspace{-8pt}
\subsection{Multi-Modal Fusion Block}\label{Fusion Block}
Effectively integrating camera, LiDAR, radar, and GPS features is nontrivial, since these modalities differ substantially in spatial structure, semantic meaning, and information density. To this end, we employ the Transformer~\cite{vaswani2017attention} to fuse heterogeneous modality features and model cross-modal dependencies within a unified representation space. At each fusion stage, the modality-specific features are first pooled and projected into a common token representation. By concatenating the tokens from all sensing modalities, we obtain the unified token sequence $\mathbf{F} \in \mathbb{R}^{\mathcal{N}\times\mathcal{D}}$, where $\mathcal{N}$ denotes the total number of tokens and $\mathcal{D}$ is the embedding dimension.
Accordingly, the query ($\mathbf{Q}$), key ($\mathbf{K}$), and value ($\mathbf{V}$) matrices are computed as~\cite{vaswani2017attention}:
\begin{equation}
\mathbf{Q}=\mathbf{F}\mathbf{W}^{\mathrm{Q}},\quad
\mathbf{K}=\mathbf{F}\mathbf{W}^{\mathrm{K}},\quad
\mathbf{V}=\mathbf{F}\mathbf{W}^{\mathrm{V}},
\label{eq:dimension for embedding}
\end{equation}
where $\mathbf{W}^{\mathrm{Q}}$, $\mathbf{W}^{\mathrm{K}}$, and $\mathbf{W}^{\mathrm{V}}$ are learnable projection matrices. The fused token representation, denoted by $\tilde{\mathbf{F}}\in\mathbb{R}^{\mathcal{N}\times\mathcal{D}}$, is then obtained through scaled dot-product attention~\cite{vaswani2017attention}:
\begin{equation}
\mathrm{Attention}(\mathbf{Q},\mathbf{K})=
\mathrm{softmax}\left(\frac{\mathbf{Q}\mathbf{K}^{\top}}{\sqrt{\mathcal{D}/h}}\right),
\end{equation}
\begin{equation}
\tilde{\mathbf{F}}=\mathrm{Attention}(\mathbf{Q},\mathbf{K})\mathbf{V},
\end{equation}
where $h$ is the number of attention heads. Through this operation, each token in $\mathbf{F}$ is updated by selectively aggregating informative features from other tokens in the shared multi-modal sequence, producing the fused representation $\tilde{\mathbf{F}}$. After fusion, $\tilde{\mathbf{F}}$ is reorganized according to the original modality partitions and returned to the corresponding branches for subsequent processing. For camera, LiDAR, and radar, the fused tokens are reshaped into feature maps and passed to the next stage. For GPS, since it does not have a native 2D spatial structure, the fused representation is kept in token form and directly forwarded to the next stage.

% Different from shallow fusion schemes that combine modality features only once or at a single representation level, BeamTransFuser adopts a four-stage hierarchical fusion design. In this way, cross-modal information can be exchanged progressively from low-level structural representations to high-level semantic features, which improves the exploitation of complementary information from heterogeneous sensing modalities for robust beam prediction.

% \vspace{-8pt}
\subsection{Beam Generator and Model Training}
After the final fusion stage, the modality features are integrated through a learnable fusion module with softmax-normalized weights (i.e., ${\color{red}\otimes}$), so that the contribution of each modality can be adaptively adjusted.
Specifically, let $\mathbf{f}_m$ denote the globally pooled feature of modality $m$, and let $\alpha_m$ and $w_m$ denote its normalized importance weight and learnable scalar score, respectively. The normalized weight $\alpha_m$ is obtained through a softmax operation:
\begin{equation}
\alpha_m=\frac{\exp(w_m)}{\sum_{j}\exp(w_j)},
\end{equation}
The final fused representation is then given by:
\begin{equation}
\hat{\mathbf{F}}=\sum_m \alpha_m \mathbf{f}_m.
\end{equation}
This fused representation is passed to a compact MLP-based beam generator to produce the beam score vector:
\begin{equation}
\hat{\mathcal{Y}} = \mathrm{MLP}(\hat{\mathbf{F}}),
\end{equation}
The predicted beam index corresponds to the beamforming vector in the codebook with the largest score. In this way, more accurate beam prediction directly translates into better beam alignment and, consequently, improved communication performance.

\section{Enhancing BeamTransFuser Robustness via Modality Generation}\label{sec:generation_only}
To improve the robustness of BeamTransFuser under incomplete sensing conditions, we introduce a generative module for missing-modality completion. The role of this module is to infer the representation of an unavailable sensing modality from the remaining observations, so that the downstream beam prediction pipeline can still operate when the multi-modal input is incomplete. Specifically, a generative model learns how to map latent variables to data samples, thereby producing outputs that follow the semantic and structural patterns observed in the training data~\cite{9555209}.

In this work, we build the proposed generation mechanism on the variational auto-encoder (VAE) framework~\cite{kingma2013auto}. In particular, this choice is motivated by the deployment requirements of the considered V2X scenario. Compared with generative adversarial networks (GANs)~\cite{goodfellow2014generative}, VAE-based models are generally easier to optimize and offer more stable training behavior, especially when the target data distribution is complex or highly diverse. Compared with diffusion-based models~\cite{croitoru2023diffusion}, VAE-based generation is also more suitable for delay-sensitive applications, since diffusion-based generation typically involves a long sequence of denoising updates at inference time, resulting in significantly higher latency. By contrast, VAE-based generation can be performed in a lightweight feed-forward manner, which is more compatible with real-time beam prediction in V2X systems~\cite{kingma2013auto}. Furthermore, since missing-modality completion requires the generated output to remain consistent with the currently observed sensing inputs, we further adopt the conditional variational auto-encoder (CVAE), in which both the encoder and decoder are conditioned on the available modalities~\cite{NIPS2015_8d55a249}.

Let $\mathbf{x}\in\mathcal{X}$ denote the available modalities and let $\mathbf{y}\in\mathcal{X}$ denote the missing modality to be generated. Under this setting, the objective is to infer $\mathbf{y}$ conditioned on $\mathbf{x}$, which is equivalent to modeling the conditional distribution $p(\mathbf{y}|\mathbf{x})$. For instance, if radar observations are missing while camera and LiDAR are available, then camera and LiDAR form $\mathbf{x}$ and the missing radar modality corresponds to $\mathbf{y}$. In this way, the generation process is guided by the observed sensing context rather than by an unconditional data prior~\cite{NIPS2015_8d55a249}. Specifically, to increase the flexibility of the conditional generation process, a latent variable $\mathbf{z}$ is introduced to capture hidden factors that are not explicitly represented by the observed modalities. Accordingly, the conditional distribution of $\mathbf{y}$ given $\mathbf{x}$ can be expressed as:
\begin{equation}
p(\mathbf{y}|\mathbf{x}) = \int_{\mathbf{z}} p_{\boldsymbol{\theta}}(\mathbf{y}|\mathbf{x},\mathbf{z})\, p(\mathbf{z}|\mathbf{x})\, d\mathbf{z},
\end{equation}
where $p_{\boldsymbol{\theta}}(\mathbf{y}|\mathbf{x},\mathbf{z})$ denotes the conditional likelihood of generating $\mathbf{y}$ given $\mathbf{x}$ and $\mathbf{z}$, parameterized by $\boldsymbol{\theta}$ (i.e., the decoder), and $p(\mathbf{z}|\mathbf{x})$ is the prior distribution of the latent variable. The generative model is then trained to learn this conditional distribution, which is equivalent to maximizing the corresponding conditional likelihood. However, since this likelihood involves marginalizing over the latent variable $\mathbf{z}$, it is generally intractable to optimize directly. To this end, a variational posterior $q_{\boldsymbol{\phi}}(\mathbf{z}|\mathbf{x},\mathbf{y})$, parameterized by $\boldsymbol{\phi}$ (i.e., the encoder), is introduced to approximate the true posterior, and the model is trained by optimizing the standard CVAE objective~\cite{NIPS2015_8d55a249,kingma2013auto}:
\begin{equation}
\begin{aligned}
\mathcal{L}_{\text{CVAE}}(\boldsymbol{\theta}, \boldsymbol{\phi})
&=
-
\mathbb{E}_{\mathbf{z} \sim q_{\boldsymbol{\phi}}(\mathbf{z}|\mathbf{x}, \mathbf{y})}
\left[
\log p_{\boldsymbol{\theta}}(\mathbf{y}|\mathbf{x}, \mathbf{z})
\right]
\\
&\quad+
\mathrm{KL}
\left(
q_{\boldsymbol{\phi}}(\mathbf{z}|\mathbf{x}, \mathbf{y})
\,\|\, p(\mathbf{z}|\mathbf{x})
\right).
\end{aligned}
\label{eq:cvae-loss}
\end{equation}
The first term in~\eqref{eq:cvae-loss} corresponds to the reconstruction objective, which encourages the decoder to produce a modality representation aligned with the target $\mathbf{y}$. The second term regularizes the variational posterior toward the latent prior, which helps improve the stability and generalization capability of the generative module. In addition, after the initial CVAE training stage, we further perform a lightweight task-aware fine-tuning step, in which the reconstruction objective is jointly optimized with a downstream beam prediction loss. This refinement improves the task relevance of the generated modality while maintaining consistency with the target feature space.

It is worth noting that explicitly modeling an input-dependent conditional prior $p(\mathbf{z}|\mathbf{x})$ generally requires an additional prior network, which increases optimization complexity as well as inference overhead. To keep the generative module lightweight for delay-sensitive V2X deployment, we follow~\cite{kingma2014semi,NIPS2015_8d55a249} and relax the conditional prior to a fixed isotropic Gaussian distribution, i.e., $p(\mathbf{z}|\mathbf{x}) = p(\mathbf{z}) = \mathcal{N}(\mathbf{0}, \mathbf{I})$.
Note that this simplification only affects the latent prior. The overall model remains conditional through the variational posterior $q_{\boldsymbol{\phi}}(\mathbf{z}|\mathbf{x},\mathbf{y})$ and the decoder $p_{\boldsymbol{\theta}}(\mathbf{y}|\mathbf{x},\mathbf{z})$. Finally, the loss in~\eqref{eq:cvae-loss} can then be optimized efficiently via stochastic gradient descent together with the reparameterization trick~\cite{NIPS2015_8d55a249}.

We illustrate the modality generation process as follows. First, the generative module is trained offline and independently of BeamTransFuser using paired incomplete and complete modality samples. After training, only the decoder is retained for inference. Given the available modalities $\mathbf{x}$, we sample a latent variable $\mathbf{z}\sim p(\mathbf{z})$ and feed both $\mathbf{x}$ and $\mathbf{z}$ into the decoder to reconstruct the missing modality. The generated modality features are then injected into BeamTransFuser at the feature level, i.e., between the feature extraction stage and the multi-modal fusion backbone in Fig.~\ref{Fig.BeamTransFuser}. Finally, the generated features are jointly processed with the observed modality features by the hierarchical fusion backbone to perform beam prediction.

\section{Performance Evaluation}\label{simultion}
\subsection{Dataset and Settings}\label{datasets}
We evaluate the proposed BeamTransFuser on the DeepSense 6G dataset~\cite{DeepSense2}, a real-world multi-modal V2X dataset that is openly accessible to the research community.
The dataset provides synchronized camera, LiDAR, radar, GPS, and beam-label data, where each label corresponds to the optimal beam index selected from a 64-beam codebook. Moreover, this dataset is well suited for evaluating the considered task, as it reflects realistic urban V2X environments with LoS/NLoS transitions, multipath propagation, traffic dynamics, and illumination variation. In particular, we focus on the urban street scenarios 31-34~\cite{DeepSense2}, in which the RSU collects multi-modal sensing data from an RGB camera, radar, LiDAR, the mmWave receiver, and GPS. After combining the development and adaptation subsets, we obtain 11,243 samples in total, among which 90\% are used for training and the remaining 10\% for validation.

For BeamTransFuser, the convolutional encoders are configured with a kernel size of $7\times7$ and stride 2, and each multi-modal fusion block uses four attention heads with a feed-forward expansion ratio of four. The model is trained for 30 epochs using focal loss with a learning rate of $10^{-4}$. In the evaluation, we adopt DBA-score and Top-$k$ accuracy as the main performance metrics~\cite{DeepSense2}. Specifically, DBA-score characterizes how far the predicted beam is from the target beam, and thus serves as a fine-grained measure of beam alignment quality~\cite{DeepSense2}. We further compare the proposed framework with the following baselines reported on the same dataset, including Avatar~\cite{a2022_deepsense}, the position-based and multi-modal methods in~\cite{10683225}, TII~\cite{tian2023multimodal}, CMDF~\cite{10.1145/3653644.3680497}, ICMFE~\cite{10912462}, and QTNs~\cite{10577431}.

\subsection{Experimental Results}
\subsubsection{Convergence Behavior and Training Time}
\begin{figure}[t]
	\centering
	\begin{subfigure}[b]{0.5\linewidth}
		\centering
		\includegraphics[width=\linewidth]{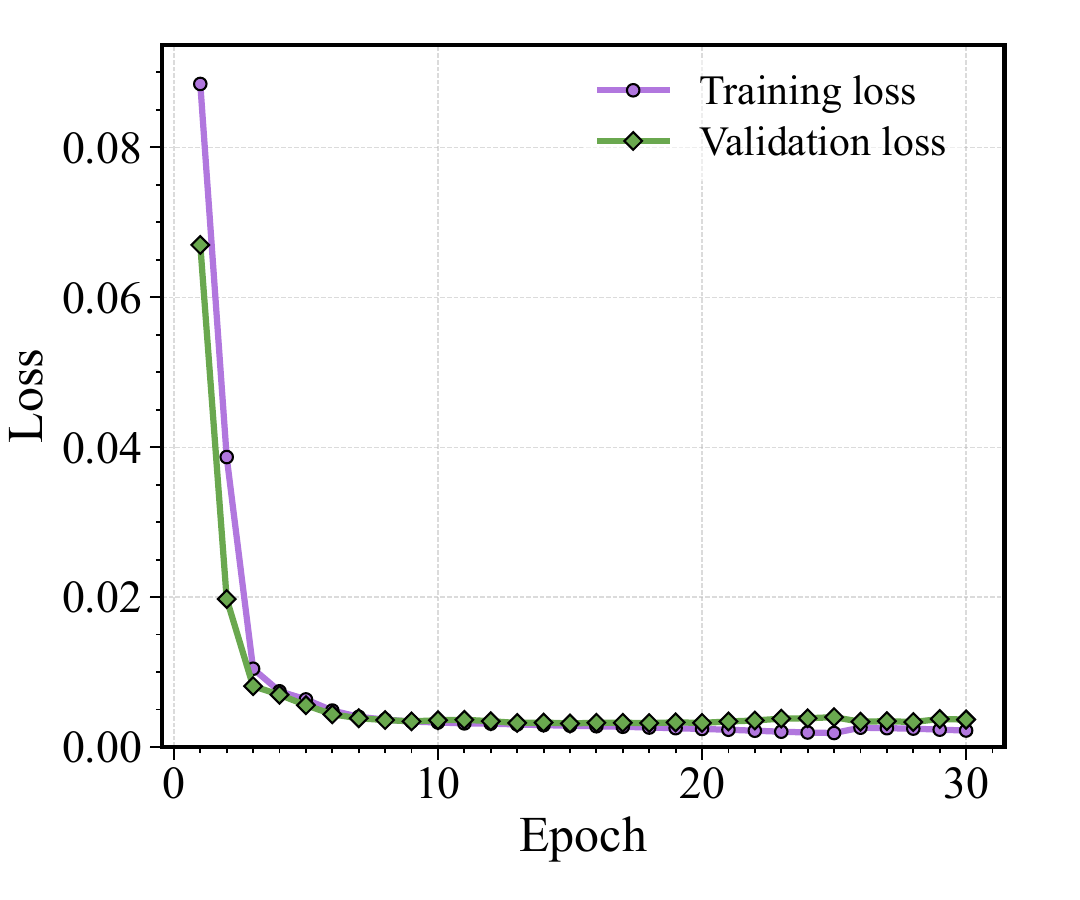}\\[-4pt]
		(a)
	\end{subfigure}%
	% \hfill
	\begin{subfigure}[b]{0.5\linewidth}
		\centering
		\includegraphics[width=\linewidth]{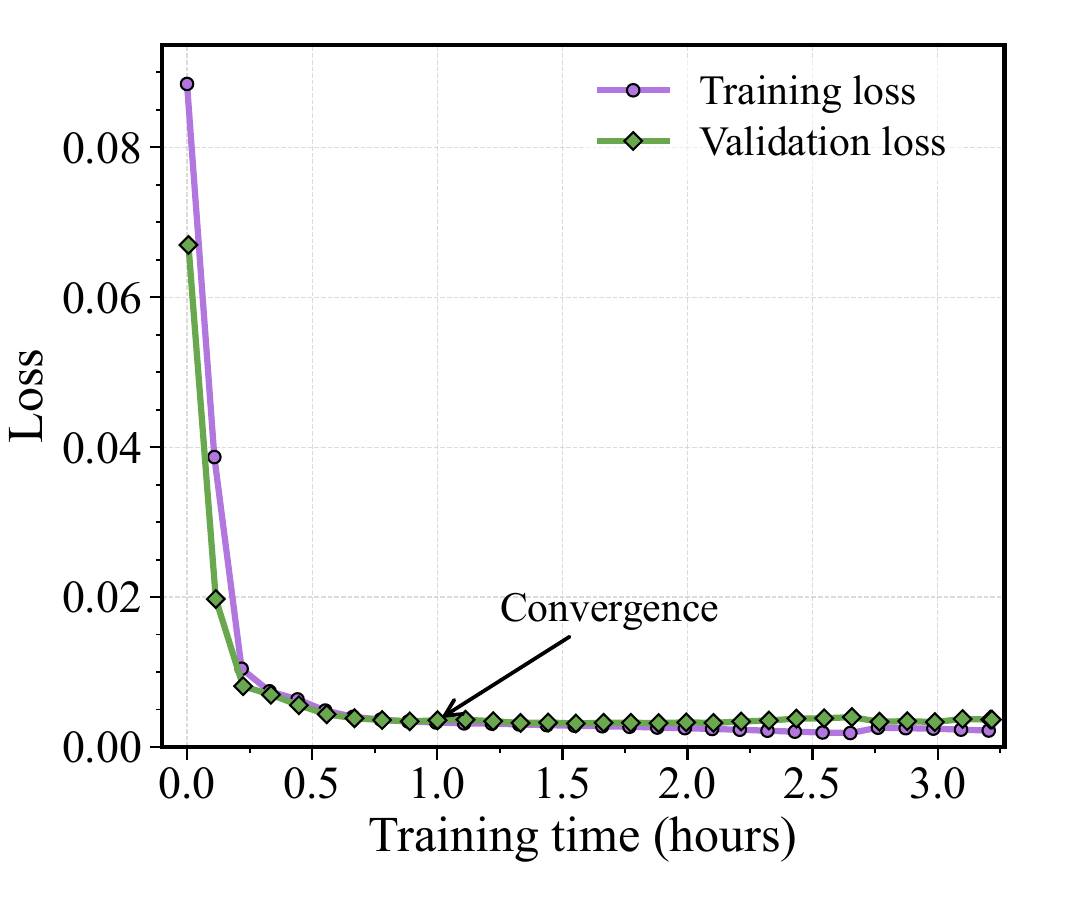}\\[-4pt]
		(b)
	\end{subfigure}
	\caption{Training behavior of the proposed model. (a) Training and validation loss versus epoch. (b) Training and validation loss versus training time.}
	\label{fig:loss}
    % \vspace{-13pt}
\end{figure}

We first examine the convergence behavior and training time in Fig.~\ref{fig:loss}. As shown in Fig.~\ref{fig:loss}(a), both the training and validation losses decrease steadily during training and gradually stabilize afterward, without any noticeable increase in the validation loss at later epochs. This indicates that the proposed model does not exhibit evident overfitting and maintains satisfactory generalization performance on the validation set. Fig.~\ref{fig:loss}(b) further presents the loss evolution with respect to training time. All training experiments were performed on an NVIDIA GeForce RTX 4090 GPU. It can be seen that the loss curves become much flatter after approximately one hour, suggesting that the optimization has largely converged, while further training only brings marginal improvement.
\subsubsection{Beam Prediction Performance}
We then evaluate beam prediction performance in terms of DBA-score. As shown in Table~\ref{tab:DBA-modal}, the proposed method achieves the highest overall DBA-score among all compared schemes. It also exhibits consistently strong results in all four scenarios, with DBA-scores of 1.0000, 0.9038, 0.8988, and 0.8945 for Scenarios 31, 32, 33, and 34, respectively. This overall stability indicates that the proposed framework can maintain reliable beam prediction under diverse urban conditions, including variations in illumination across daytime scenarios (Scenarios 31 and 32) and nighttime scenarios (Scenarios 33 and 34), as well as different propagation conditions under both LoS and NLoS settings.

\vspace*{0.04in}
\begin{table}[t]
\centering
\caption{DBA-score results of selected beam prediction schemes}
\label{tab:DBA-modal}
\begin{threeparttable}
\renewcommand{\arraystretch}{0.9}
\resizebox{\linewidth}{!}{
\begin{tabular}{cccccc}
    \toprule
    \textbf{Scheme} & \textbf{Overall} & \textbf{S31} & \textbf{S32} & \textbf{S33} & \textbf{S34} \\
    \midrule
    Avatar~\cite{a2022_deepsense} & 0.7162 & 0.6536 & 0.7074 & 0.8576 & 0.7120 \\
    \cite{10683225} & -- & -- & 0.8906 & -- & -- \\
    TII~\cite{tian2023multimodal} & 0.7844 & 0.7298 & 0.7852 & 0.8462 & 0.8433 \\
    CMDF~\cite{10.1145/3653644.3680497} & 0.8910 & -- & -- & -- & -- \\
    ICMFE~\cite{10912462} & 0.8969 & 1.0000 & 0.9020 & 0.8874 & 0.9074 \\
    QTNs~\cite{10577431} & -- & 0.7605 & 0.8707 & 0.8864 & 0.9124 \\
    \textbf{BeamTransFuser (Ours)} & \textbf{0.9129} & \textbf{1.0000} & \textbf{0.9038} & \textbf{0.8988} & \textbf{0.8945} \\
    \bottomrule
\end{tabular}
}
\begin{tablenotes}
    \footnotesize
    \item -- indicates an unreported result in the cited work.
\end{tablenotes}
\end{threeparttable}
\end{table}

However, some baselines follow a different trend. For instance, Avatar~\cite{a2022_deepsense} and TII~\cite{tian2023multimodal} both achieve higher DBA-scores in Scenario 33 than in Scenario 32, with improvements of 0.15 and 0.061, respectively. This may indicate that some nighttime scenes, such as those with weaker background clutter or more homogeneous illumination, are more favorable to these methods. At the same time, the amount of improvement is not uniform across baselines, which suggests that their effectiveness is strongly influenced by how different sensing modalities are utilized and fused.

Compared with these baselines, BeamTransFuser remains much more stable from Scenarios 32 to 34, reflecting stronger robustness to environmental variation. It is also worth noting that, although QTNs~\cite{10577431} and the method in~\cite{10912462} obtain slightly higher DBA-scores in Scenario 34, the margins are small, namely 1.99\% and 1.42\%, respectively. From an overall perspective, BeamTransFuser still ranks first with a DBA-score of 0.9129. These results suggest that, even though some baselines perform competitively in individual scenarios, BeamTransFuser remains more stable over the full set of evaluated conditions, demonstrating stronger generalization and more reliable beam prediction for realistic V2X deployment.

The strong DBA performance of BeamTransFuser can be attributed to two main factors. First, the model jointly exploits complementary information from camera, LiDAR, radar, and GPS, allowing geometric structure, visual semantics, motion cues, and location information to be utilized in a unified manner. This is particularly beneficial in urban V2X scenarios, where a single sensing modality may become unreliable under blockage, NLoS propagation, or environmental variation. Second, the hierarchical Transformer-based fusion design enables cross-modal information to be progressively integrated across multiple representation levels, rather than being merged only once at the final stage. As a result, BeamTransFuser can better capture modality-specific structural information together with semantic context, which leads to more robust and consistent beam prediction across diverse scenarios.

\subsubsection{Performance with Modality Completion}
We next investigate the robustness of BeamTransFuser in the presence of missing modalities. To emulate practical cases where one sensing modality is unavailable, e.g., due to sensor failure or incomplete deployment, we replace the missing input with either zero-filled data or Gaussian noise and treat the resulting outputs as degraded baselines. We then compare these baselines with the case where the missing modality is reconstructed by the proposed generative model.

% $\binom{2^b}{k}$

As shown in Table~\ref{tab:missing_modality_single}, when Radar or LiDAR is unavailable, the Top-1 accuracy drops sharply under both zero-filled and Gaussian-noise replacements, indicating that the absence of these sensing inputs severely degrades beam prediction performance. By contrast, replacing the missing modality with CVAE-generated features substantially restores the prediction accuracy, which verifies the effectiveness of the proposed modality generation mechanism under incomplete sensing conditions. Specifically, in the missing-Radar case, the Top-1 accuracy improves from 6.67\% to 45.51\%, while in the missing-LiDAR case, it increases from 6.67\% to 56.18\%. These notable gains indicate that the generative module can recover informative modality features from the remaining observations and provide meaningful compensation when one sensing source is unavailable. We do not report the generated-camera case, because reconstructing high-dimensional RGB features is considerably more difficult and introduces much higher complexity. Even so, the strong recovery achieved in the missing-Radar and missing-LiDAR settings already demonstrates the practical value of the proposed generation mechanism for robust beam prediction under incomplete sensing.

\vspace*{0.04in}
\begin{table}[t]
\centering
\caption{Top-\(k\) beam prediction accuracy under missing-modality conditions}
\label{tab:missing_modality_single}
\footnotesize
\setlength{\tabcolsep}{3pt}
\renewcommand{\arraystretch}{1.05}
\begin{tabular}{c c c c c}
    \toprule
    \textbf{Miss. Mod.} & \textbf{Repl. Cond.} & \textbf{Top-1 (\%)} & \textbf{Top-2 (\%)} & \textbf{Top-3 (\%)} \\
    \midrule
    \multirow{3}{*}{Radar}
      & Zero-filled      & 6.67  & 9.07  & 15.02 \\
      & Gaussian noise   & 6.67  & 9.42  & 14.22 \\
      & Gen (Cam+LiDAR)  & \textbf{45.51} & \textbf{70.31} & \textbf{82.58} \\
    \midrule
    \multirow{3}{*}{LiDAR}
      & Zero-filled      & 6.67  & 9.07  & 13.96 \\
      & Gaussian noise   & 6.67  & 9.07  & 13.78 \\
      & Gen (Cam+Radar)  & \textbf{56.18} & \textbf{79.82} & \textbf{89.07} \\
    \midrule
    \multirow{2}{*}{Camera}
      & Zero-filled      & 2.40  & 4.27  & 6.40  \\
      & Gaussian noise   & 3.02  & 6.31  & 8.27  \\
    \bottomrule
\end{tabular}
% \vspace{-6pt}
\end{table}

\vspace{-10pt}
\section{Conclusion}\label{conclusion}
\vspace{-10pt}
In this work, we have proposed a multi-modal beam prediction framework for V2X networks based on hierarchical fusion and modality generation. Specifically, we have developed BeamTransFuser, which exploits complementary information from camera, LiDAR, radar, and GPS through progressive Transformer-based fusion for accurate and robust beam prediction. To further handle possible incomplete sensing conditions in practical deployment, we have incorporated a generative module that reconstructs missing modality features from the available observations, thereby improving robustness without requiring retraining. Experimental results on a real-world multi-modal V2X dataset have demonstrated the effectiveness of the proposed approach in improving beam prediction accuracy over competing schemes. They also confirm that the generative module enables stable beam prediction when part of the sensing input is unavailable. In future work, we will extend this framework to more challenging urban scenarios and multi-vehicle settings.
\vspace{-5pt}
\bibliographystyle{IEEEtran}
\bibliography{bibRef}
\vspace{-10pt}

\end{document}